\newif\ifdraft
 \draftfalse

\documentclass[11pt]{article}

\usepackage[letterpaper,margin=1in]{geometry}
\usepackage[parfill]{parskip}

\usepackage{authblk}
\usepackage[toc,page,header]{appendix}
\usepackage[utf8]{inputenc} % allow utf-8 input
\usepackage[T1]{fontenc}    % use 8-bit T1 fonts
\usepackage[colorlinks,citecolor=blue,urlcolor=blue,linkcolor=blue,linktocpage=true]{hyperref}       % hyperlinks
\usepackage{url}            % simple URL typesetting
\usepackage{booktabs}       % professional-quality tables
\usepackage{amsfonts}       % blackboard math symbols
\usepackage{nicefrac}       % compact symbols for 1/2, etc.
\usepackage{microtype}      % microtypography
\usepackage{xcolor}         % colors
\label{key}

\usepackage{comment}
\usepackage{amsmath,amssymb,amsfonts}
\usepackage{xspace}

\usepackage{epsfig}
\usepackage{graphicx}
\usepackage{wrapfig}
\usepackage{subfigure}
\usepackage{multirow}
\usepackage{hyperref}
\usepackage{graphicx}
\usepackage{caption}
\usepackage{array}
\usepackage{bbm}
\usepackage{color}
\usepackage{enumerate}
\usepackage{enumitem}
\usepackage{mathtools}
\usepackage{amsmath,amssymb,amsthm,bm}
\usepackage{thmtools}
\usepackage{amsfonts,graphicx}
\usepackage{mathrsfs}
\usepackage{amsmath,amssymb,amsfonts}
\usepackage{textcomp}
\usepackage{bbold}
\usepackage{graphicx}
\usepackage{textcomp}
\usepackage{xcolor}
\usepackage[authoryear]{natbib}

\author[1]{Yingyan~Zeng}
\affil[1]{University of Cincinnati}
\author[2]{Ismini~Lourentzou}
\affil[2]{University of Illinois}
\author[3]{Xinwei~Deng}
\affil[3]{Virginia Tech}
\author[3]{Ran~Jin}

\providecommand{\keywords}[1]{\textbf{\textit{Keywords:}} #1}
\newcommand*{\eg}{\emph{e.g.},\@\xspace}
\newcommand*{\ie}{\emph{i.e.},\@\xspace}
\newcommand*{\etc}{\emph{etc.}\@\xspace}
\usepackage{bm}

\title{FAIR: Facilitating Artificial Intelligence Resilience in Manufacturing Industrial Internet}
\begin{document}

%\doparttoc % Tell to minitoc to generate a toc for the parts
%\faketableofcontents % Run a fake tableofcontents command for the partocs

% \part{} % Start the document part
% \parttoc % Insert the document TOC

\maketitle

\begin{abstract}
Artificial intelligence (AI) systems have been increasingly adopted in the Manufacturing Industrial Internet (MII).
Investigating and enabling the AI resilience is very important to alleviate profound impact of AI system failures in manufacturing and Industrial Internet of Things (IIoT) operations, leading to critical decision making.
However, there is a wide knowledge gap in defining the resilience of AI systems and analyzing potential root causes and corresponding mitigation strategies. 
In this work, we propose a novel framework for investigating the resilience of AI performance over time under hazard factors in data quality, AI pipelines, and the cyber-physical layer. The proposed method can facilitate effective diagnosis and mitigation strategies to recover AI performance based on a multimodal multi-head self latent attention model. 
The merits of the proposed method are elaborated using an MII testbed of connected Aerosol\textsuperscript{\textcircled{\tiny R}} Jet Printing (AJP) machines, fog nodes, and Cloud with inference tasks via AI pipelines. 
\end{abstract}


\section{Introduction} \label{Sec: Intro}

% computation services in MII
A Manufacturing Industrial Internet (MII) connects manufacturing equipment, physical processes, systems, and networks via ubiquitous sensors, actuators, and computing units \citep{chen2018predictive}. 
By enabling the seamless collection of high-speed, large-volume data, MII establishes the digital foundation necessary for deploying Artificial Intelligence (AI) models that deliver critical computational services, such as quality modeling, process variation analysis, fault prognosis and diagnosis, and optimization \citep{arinez2020artificial}, which enhance manufacturing efficiency, reduce operational costs, and enable intelligent automation.

% Why important
As AI becomes increasingly integrated into manufacturing decision-making within MII, ensuring the resilience of AI systems is critical for maintaining reliable and continuous computation services \citep{chen2020adapipe}.
In the context of MII, an AI system consists of three interconnected layers: (i) the data layer, representing the collected manufacturing data; (ii) the AI pipeline layer, where data are processed via the deployed AI models and pipeline ranking systems \citep{chen2020adapipe}; (iii) the cyber-physical layer, encompassing the Fog-Cloud computing infrastructure that supports the communication and computation in the system \citep{wang2020monitoring, chen2024lori}.
Thereafter, AI systems face challenges from three primary types of root causes, each tied to a critical system component: (i) data quality issues; (ii) AI model singularity; and (iii) cyber-physical layer failures.
% First, data quality issues stem from abrupt data distribution shifts, sensor degradation, and changing operation conditions.
% Second, AI model singularity occurs when the model's parameters or mathematical structure become degenerate, resulting in issues such as ill-conditioned optimization and vanishing or exploding gradients.
% Finally, we focus on the cyber-physical layer failures including communication failures caused by cyber attacks or hardware malfunctions due to edge node degradation. 
These hazards can lead to severe AI modeling performance degradation, resulting in incorrect predictions, faulty decisions, and economic losses. 
Therefore, it is necessary to develop a comprehensive framework that formally defines AI resilience in MII, quantifies the resilience performance, and proposes mechanisms for detecting and mitigating hazards in AI systems.

% literature - existing approach - knowledge gap (under AI in MII setting) - quantitative way to define the resilience of AI systems
In the literature, system resilience is generally defined as a system's ability to withstand, respond to, and recover from unexpected disruptions (\ie hazards) \citep{poulin2021infrastructure}. 
To evaluate the resilience of general infrastructure systems designed to provide continuous services \citep{hall2016introducing}, resilience curves have been widely studied to derive magnitude-based, duration-based, rate-based, and threshold-based metrics \citep{ poulin2021infrastructure, cheng2023reliability}.
In the context of MII, a closely related domain is Internet-of-Things (IoT) systems, where resilience is linked with attributes such as confidentiality, integrity, reliability, maintainability, and safety \citep{berger2021survey}. 
% Various resilience metrics, including fault-tolerance coverage and degree of replication, have been proposed to capture the multi-dimensional nature of IoT resilience \citep{avizienis2004basic}. 
While existing frameworks provide a foundation for understanding AI system resilience in MII, they lack an approach that addresses the unique challenges of AI systems under the influence of three key interconnected layers.
Despite the importance of resilience, most AI evaluation frameworks remain narrowly anchored to static performance metrics (\eg accuracy, precision) \citep{flach2019performance}.
These metrics, while useful for offline model validation, provide limited insight into how AI systems degrade, adapt, or recover when confronted with heterogeneous disruptions in an online computation service in MII.
Compounding this issue is the absence of quantitative resilience metrics for AI systems in MII.
Furthermore, limited studies substantiate these frameworks with empirical demonstrations and experimental validation, hindering further diagnosis and mitigation efforts.
% challenge on the modeling with high-dimensional Multimodal signals

The objective of this work is to create a novel framework of resilient AI in MII which aims to (i) create quantitative metrics to evaluate the resilience of the AI system; (ii) diagnose root causes of AI system performance degradation; and (iii) automate context-aware mitigation strategies for the failures. 
The proposed framework will establish a solid foundation for understanding, quantifying, and improving the resilience of the AI systems in MII.

To achieve the objective, we first identify and categorize the different types of root causes that affect each of the three layers of an AI system in MII.
Building on the taxonomy of failures and informed by the literature on infrastructure resilience, we consider two new metrics: temporal resilience and performance resilience to jointly quantify the AI system’s capacity to absorb and recover from hazards arising from the identified root causes.
Moreover, the proposed framework focuses on detecting failures and identifying their root causes to enable timely mitigation strategies that enhance system stability.
It is known that the MII’s integrated digital infrastructure can enable a seamless collection of runtime metrics of computation nodes, including CPU and memory utilization, download and upload bandwidths, along with performance from AI pipeline. 
Such information provides granular visibility of the system, which serves as a ``side channel" to detect and diagnose hazards.
However, the high-dimensional multimodal runtime data pose significant challenges for constructing proper diagnosis models, where multiple root causes can occur simultaneously. 
To address these challenges, we propose a Multimodal Multi-head Self Latent Attention  (MMSLA) model to accurately diagnose root causes by capturing the dependencies between latent features associated with different root causes.
Finally, mitigation strategies for different failure scenarios are developed based on several promising approaches from literature. 
The effectiveness of the resilience metrics, MMSLA model, and mitigation actions is evaluated within an MII environment comprising Cloud computing and fog nodes. 
%The MMSLA model is benchmarked against multiple existing methods and demonstrates superior performance.

% To achieve this goal, we first identify the different types of root causes in detail for each of the three layers in AI system in MII and the corresponding qualitative impact on the system performance.
% Based on the impact in MII and referring to the literature on infrastructure resilience, we define the temporal and performance resilience of AI models, which quantifies the ability of the AI system to absorb the hazards caused by the identified root causes.
% Furthermore, it is important to detect the hazard and identify the root cause so that mitigation can be made to improve the resilience of the system.
% It is nontrivial to diagnose accurately from various root casues which may happen simultaneously and also with high-dimensional multi-model signal collected from MII.
% To address these challenges, we propose a Multi-Head Self-Latent Attention Model (MMSLA) to monitor the system resilience and diagnosis the root causes.
% The proposed MMSLA model effectively learns the correlation between latent features for multiple root causes simultaneously, improving the diagnosis accuracy.
% Finally, the mitigation for different causes are proposed by adopting the approaches from literature.
% The effectiveness of the designed resilience metrics, proposed MMSLA model and the corresponding mitigation actions was evaluated in a MII which contains cloud, computing edge nodes. 
% The HMMSLA model was compared with multiple benchmarks and demonstrate superior performance. 

The remainder of this paper is organized as follows.
Section~\ref{Sec: methodology} introduces the proposed resilient AI framework. 
Section~\ref{Sec: CaseStudy} evaluates the performance of the framework with a comprehensive MII case study for Aerosol Jet® Printing (AJP) process quality modeling.
We conclude the work with some discussion of future work in Section~\ref{Sec:conclusion}.

% Recovery: the time to recover to a desired performance level depends on the severity of damage, the repair rate, and resources.
% different performance metrics for cyber-physical layer

% # Main contributions
% design the experiments to simulate the resilience under three root causes
% propose model to diagnosis
% predefined mitigation strategies - 1/2/3  (adopt from literature?)

% resilience of AI systems (three types of root causes: data quality, pipeline singularity, cyber failures - potential issues)
% AB together, B is embedded factor - for the experimental results

% Why it is important

% # cbeal for the data quality as a mitigation approach

\section{Methodology}\label{Sec: methodology}
\subsection{The AI System and Resilience Metrics In MII}
\begin{figure}[ht]
    \centering
    \includegraphics[width=0.9\linewidth]{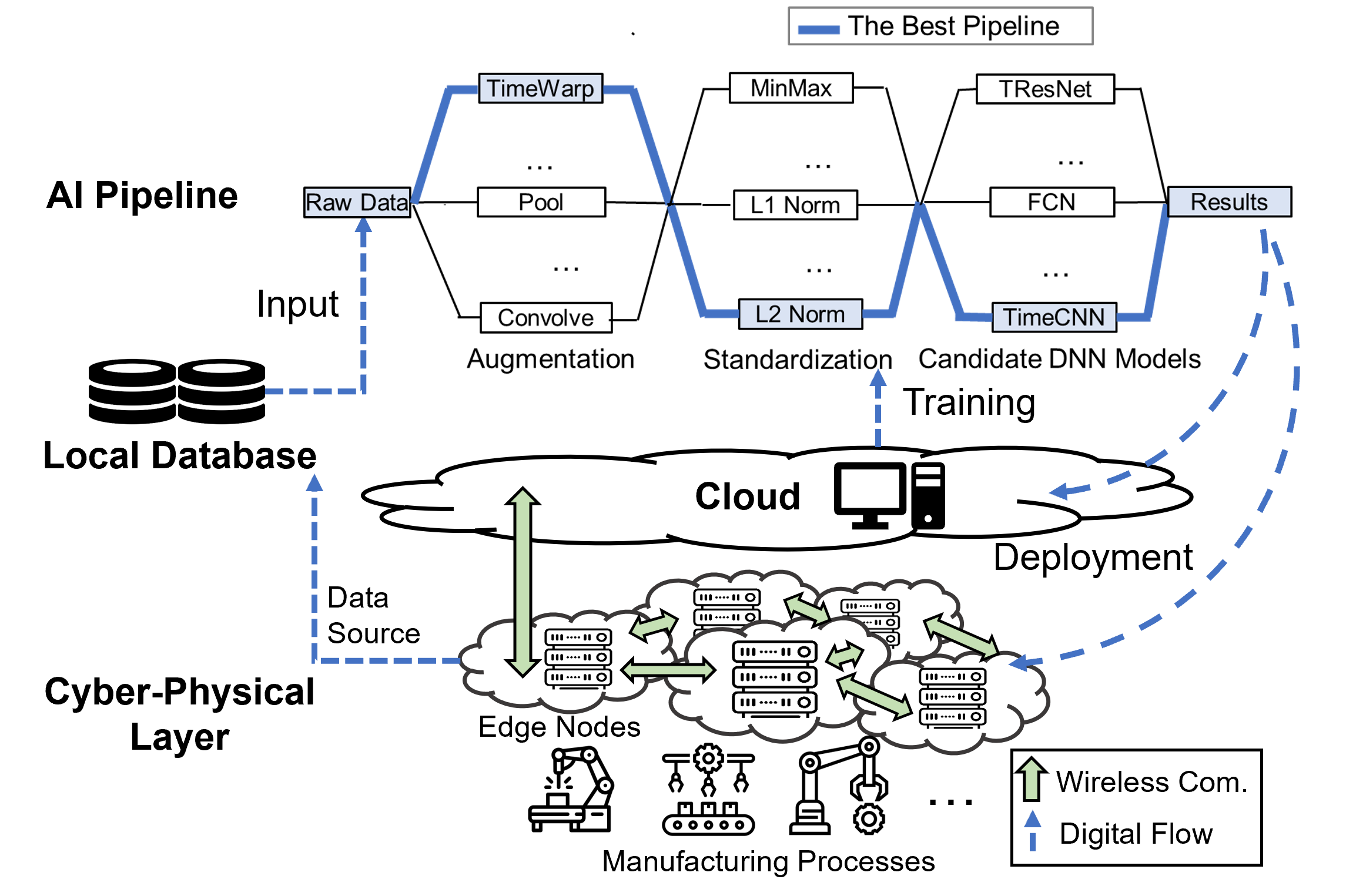}
    \caption{The AI system in Manufacturing Industrial Internet }
    \label{fig:AI_sys}
\end{figure}

% # define the AI system
The AI system in MII consists of three layers: data layer, AI pipeline layer, and cyber-physical layer. 
In this study, we focus on the AI system that provides supervised learning-based computation services (\eg quality modeling).
As shown in Fig.~\ref{fig:AI_sys},  sensor data collected via MII during manufacturing processes are stored in the local database.
The collected data serves as input to the AI pipeline, which processes it through a sequence of steps to support online decision-making. 
Specifically, we focus on multivariate time series classification pipelines, which consist of data augmentation, standardization, and Deep Neural Network (DNN) classifiers \citep{shojaee2021deep}.
The AI pipelines are ranked and trained in the Cloud using historical data and then deployed in the cyber-physical network for real-time inference.
The Cloud functions both as an orchestrator, assigning computation tasks to nodes, and as a computation node, executing tasks as needed.
The fog nodes (\eg digital signal processors and GPUs) are positioned close to the machines for efficient processing with low communication latency. 
They execute the deployed AI pipelines and return results to the Cloud for further analysis and decision-making.

% # define the current fatcors  - and their impact on the AI system performance
The AI system's performance is measured by metrics, including accuracy, precision, F1 score of each supervised learning task over the operation time, which should remain within an expected range under normal conditions.
To quantify the system's resilience under hazards, we identify key root causes from each layer that may impact performance.
In the data layer, performance degradation can arise from (i) sensor contamination or failure, resulting in inaccurate measurements or missing data in several modalities \citep{liu2020missing};
(ii) sensor degradation and calibration issues may cause a low signal-to-noise ratio (SNR), which makes it difficult for AI systems to distinguish meaningful variations from noise \citep{montgomery2009statistical};
(iii) data distribution shifts in input variables occur due to manufacturing customization, where process settings are adjusted to accommodate different product specifications, potentially leading to time varying data distributions \citep{li2022cluster};
(iv) imbalanced class distribution in the data, where data from conforming processes significantly outweighs nonconforming samples \citep{liu2022defect, zeng2023ensemble, zeng2023synthetic}.
Within the AI pipeline layer, a critical hazard arises from structural instability in model architectures, which we refer to as model singularity. 
The model's parameters or mathematical structure become degenerate, resulting in issues such as vanishing/exploding gradients during optimization \citep{tan2019vanishing}, which manifest as erratic predictions (\eg NaN values or extreme outliers) during inference \citep{hanin2018neural}.
In the cyber-physical layer, we consider two primary hazards: (i) communication channel disruptions, triggered by cyber-attacks such as distributed denial-of-service (DDoS) intrusions \citep{lo2010cooperative}, and (ii) fog node failures, caused by hardware degradation and failure (\eg overheating, memory leaks) or software-induced resource exhaustion. 
These hazards degrade real-time inference capabilities, manifesting as latency spikes, data packet loss, or unplanned downtime.
In summary, hazards across the three layers degrade AI system performance in MII, manifesting in distinct ways: data-layer disruptions corrupt input integrity, pipeline-layer instabilities induce erratic predictions, and cyber-physical failures cripple real-time inference.

\begin{figure}[ht]
    \centering
    \includegraphics[width=0.6\linewidth]{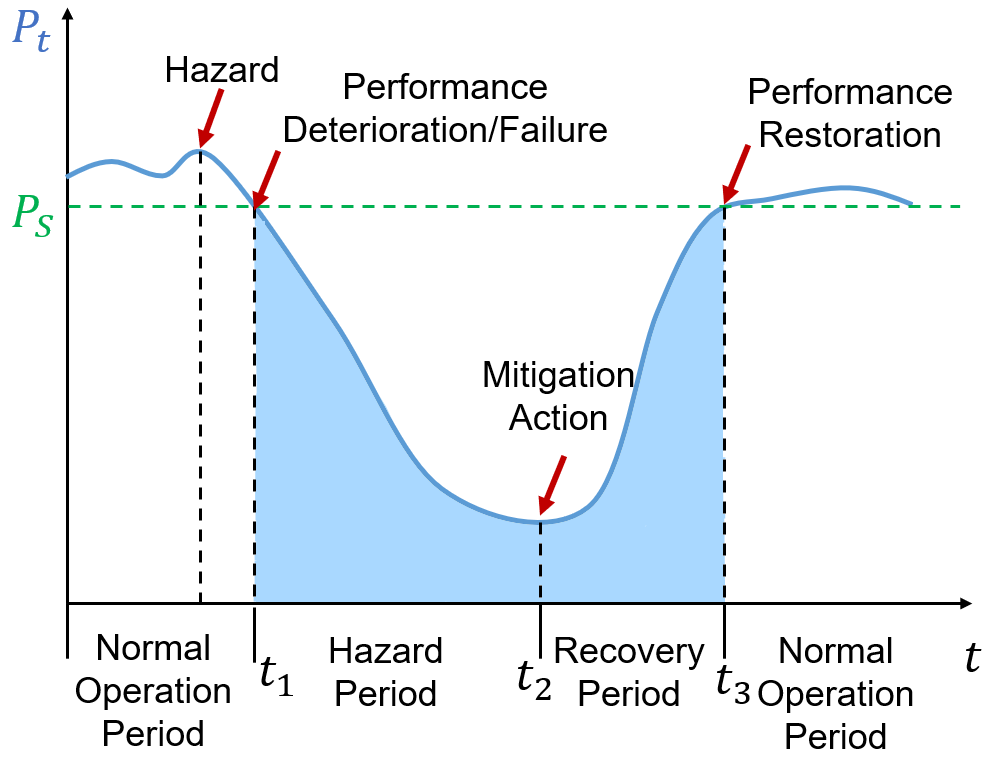}
    \caption{AI system subject to performance deterioration and recovery actions in MII.}
    \label{fig:metrics}
\end{figure}

To evaluate the resilience of the AI systems against such hazards, we propose two metrics: temporal resilience and performance resilience to quantify the impact and how the AI system recovers from the failures. 
Inspired by \citep{cheng2023reliability}, Fig.~\ref{fig:metrics} shows an example of the performance curve of an AI system during its operation. 
Let $P_t$ denote the performance of the AI system at time $t$ (\eg classification accuracy for quality modeling), and $P_S$ represent the minimum satisfactory threshold  in the computation service.
Under normal operation, $P_t \geq P_S$.
When a harzard factor is applied to the AI system, the hazard-triggered failure occurs at $t_1$ when $P_t < P_S$.
After detecting the failure and diagnosing the root cause, mitigation begins at $t_2$, which restores the system's performance to the desired level $P_S$ by $t_3$.
% Quantitative definition
\begin{table}[t!] 
\centering
\caption{The proposed resilience metrics.}
\resizebox{0.95\linewidth}{!}{
\begin{tabular}{llll}
\toprule
Category & Metrics & Expression & Definition \\\midrule
\multirow{2}{*}{\shortstack[l]{Temporal \\Resilience}} & \shortstack[l]{Failure \\ Duration (FD)} & $t_3-t_1 \downarrow$ & \shortstack[l]{Duration of Hazard Period \\ and Recovery Period}\\
 &  \shortstack[l]{Recovery \\ Efficiency (RE)}  & $(t_3-t_2)/(t_3-t_1) \uparrow$  & \shortstack[l]{Ratio of Recovery Time to \\Total Failure Duration} \\\hline
\multirow{2}{*}{\shortstack[l]{Performance \\Resilience }} & \shortstack[l]{Performance \\ Retention (PR)} & $\int_{t_1}^{t_3}P_t/(t_3-t_1) \uparrow$& \shortstack[l]{Average Performance\\ during Failure} \\
 & \shortstack[l]{Restoration \\Rate (RR)} & $\int_{t_3}^{t_3+\Delta t}P_t/\int_{t_1}^{t_1-\Delta t}P_t \uparrow$ & \shortstack[l]{Restored Performance Relative \\to Pre-failure Baseline} \\
 \bottomrule
\end{tabular}}\label{tab:res_metrics}
\end{table}

% add reference - larger better, or something

As detailed in Table.~\ref{tab:res_metrics}, temporal resilience quantifies recovery speed through two dimensions:
the failure duration as the total downtime from hazard onset to full recovery (shorter is better), and recovery efficiency, which reflects how effectively the system restores operation (higher is better).
Performance resilience, on the other hand, assesses decision-making robustness of the system via the performance retention and performance restoration rate (both higher is better).
Together, these metrics assess how AI systems withstand, adapt to, and recover from hazards comprehensively.
% Together, these metrics provide a comprehensive evaluation of how AI systems withstand, adapt to, and recover from hazards.

\subsection{A Multimodal Multi-head Self Latent Attention Model for Root Cause Diagnosis}

\begin{figure}[ht]
    \centering
    \includegraphics[width=0.8\linewidth]{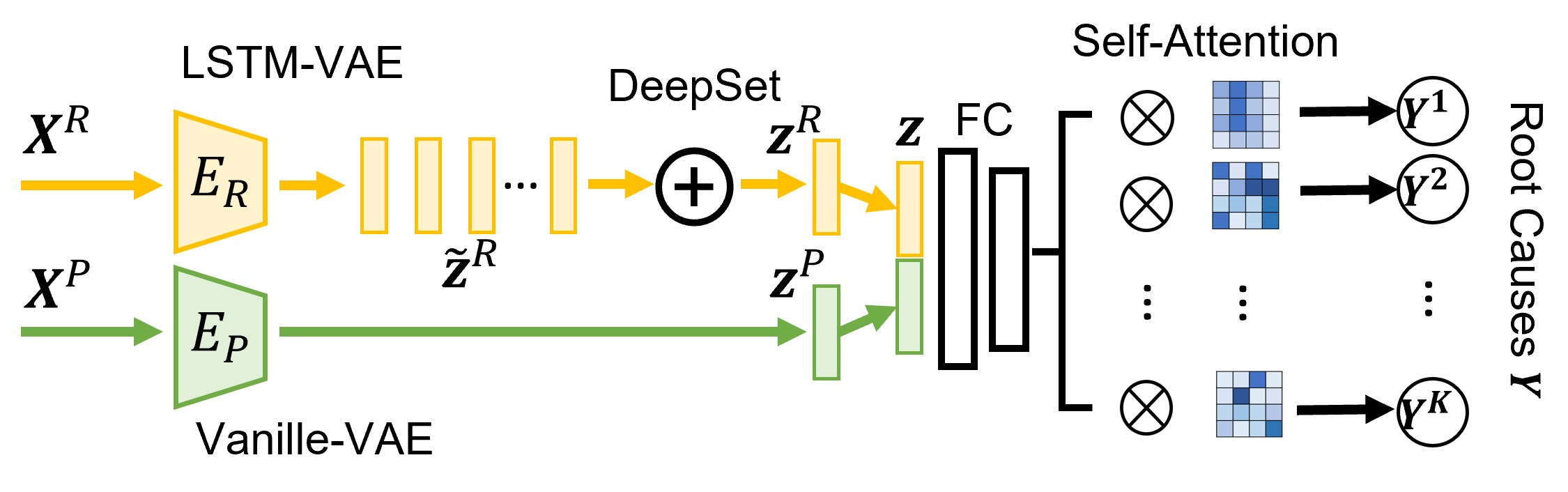}
    \caption{The proposed MMSLA model}
    \label{fig:model}
\end{figure}

Achieving high resilience in AI systems hinges on precise root-cause diagnosis across data, model, and cyber-physical layers, ensuring correct mitigation actions align with the underlying failure mode.
% This closed-loop resilience framework enables the AI system to adapt to disruptions,  sustaining operational continuity in dynamic manufacturing environments.
% The MII’s integrated digital infrastructure enables a seamless collection of runtime metrics of the computation nodes, such as CPU utilization, memory utilization, download and upload bandwidths, along with performance  from supervised learning tasks. 
% This information provides granular visibility of the system's performance.
% However, the high-dimensional multimodal data pose significant challenges to the diagnosis model due to their variable dimensions across computation tasks, and the simultaneous occurrence of multiple root causes.
To provide accurate diagnosis across multiple root causes based on multimodal data with varying and high dimensions, we propose the MMSLA model (Fig.~\ref{fig:model}) to predict the root cause types based on runtime metrics and performance of the computation task as input. 
The MMSLA encodes the input data from multiple modalities into a lower-dimensional latent space and employs a set function-based neural network to standardize varying dimensions across samples.
Multi-head attention is then applied to the latent variables, with each head focusing on a specific factor, enabling the model to simultaneously learn distinct correlations between latent variables for each root cause.

In the AI system in MII, we assume the computation service is provided for a batch of data each time.
After completing a computation task $i$, the supervised learning performance $\bm{X}^{P}_i \in \mathbb{R}^p$  is sent to the Cloud, where $P$ represents the performance metrics' dimension.
Meanwhile, the assigned computation node collects runtime metrics as $\bm{X}^{R}_i \in \mathbb{R}^{n_i \times R}$, where $n_i$ is the number of timestamps for task $i$, and $R$ is the number of monitored runtime metrics.

To fuse the information from both modalities, the MMSLA model first encodes $\bm{X}^{P}_i$ and $\bm{X}^{R}_i$ by an LSTM-Variational Autoencoder (VAE) $E_R(\cdot)$~\citep{wang2017time} and a vanilla VAE $E_P(\cdot)$~\citep{kingma2013auto}, respectively. 
Let $n$ be the sample size, \ie the number of executed computation tasks.
$E_P(\cdot)$ encodes $\bm{X}^{P}$ to a latent variable $\bm{z}^P \in \mathbb{R}^{n\times u^P}$, where  $u^P$ is the latent dimension.
Similarly, $E_R(\cdot)$ encodes $\bm{X}^{R}$ to $\tilde{\bm{z}}^R \in \mathbb{R}^{n\times n_i \times u^R}$ with $u^R$ as the latent dimension.
During the process, the VAE loss is employed on each modality, which minimizes the Kullback-Leibler (KL) divergence:
\begin{align}
\mathcal{L}^{VAE} & =  -  \mathbb{E}_{q_\phi(\bm{z}^P \mid \bm{x}^P)}\left[\log p_\theta\left(\bm{X}^{P} \mid \bm{z}^{P}\right)  \right] \nonumber \\
& -\mathbb{E}_{q_\phi(\tilde{\bm{z}}^{R} \mid \bm{x}^R)}\left[\log p_\theta\left(\bm{X}^{R} \mid \tilde{\bm{z}}^{R}\right) \right] \nonumber \\
& +\alpha D_{KL}\left(q_\phi(\bm{z}^{P} \mid \bm{X}^{P}) \| p_\theta(\bm{z}^{P})\right) \nonumber\\ 
& +\alpha D_{KL}\left(q_\phi(\tilde{\bm{z}}^{R} \mid \bm{X}^{R}) \| p_\theta(\tilde{\bm{z}}^{R})\right), 
\end{align}
where $q_\phi(\bm{z}|\bm{x})$ approximates the true latent distribution $p_{\theta}(\bm{z})$, and $\alpha \ge 0$ is the weight for KL loss.
Here the same weight is used for each modality to encourage a balanced latent representation.

To standardize the varying dimension $n_i$, a canonical model architecture, DeepSet \citep{zaheer2017deep}, is applied to $\tilde{\bm{z}}^R$, mapping it to $\bm{z}^{R} \in \mathbb{R}^{n\times u^R}$.
In particular, the DeepSet model is  a set function $f(S)=\rho(\sum_{\tilde{\bm{z}}^R\in S}\phi(\tilde{\bm{z}}^R))$, where both $\rho$ and $\phi$ are neural networks.
Hereby, the latent vectors $\bm{z}^{P}$ and  $\bm{z}^{R}$ are concatenated as latent features $\bm{Z}$.
To capture root cause-specific feature importance and inter-latent variable dependencies, we propose to use a multi-head self attention mechanism, where each head focuses on the dependencies associated with a specific root cause. 
This approach jointly models latent variables across all root causes, providing greater efficiency than developing separate diagnosis models for each root cause. 
Additionally, by learning from shared latent representations, the model effectively captures correlations between root causes, enhancing diagnostic accuracy.

Let $K$ be the number of all root causes of different types.
For root cause $j$, the following layer is applied to the latent features:
\begin{align}
        f^j_{\mathrm{att}}(\bm{Z}) = \bm{Z} \odot \operatorname{softmax}\left(\bm{W}_{f_{\mathrm{att}}}^j \bm{Z}+\bm{b}_{f_{\mathrm{att}}}^j\right),
\end{align}
where the weight matrix $\bm{W}_{f_{\mathrm{att}}}^j \in \mathbb{R}^{(u^P+u^R)\times(u^P+u^R)}, j\in K$ represents the relations between the input latent variables for root cause $j$, $\bm{b}_{f_{\mathrm{att}}}^j$ is the corresponding bias, and $\odot$ refers to the Hadamard product.
Afterward, the output of this attention layer is used as the input for the following two fully connected layers to predict the root cause class label $\bm{Y}^j$:
\begin{align}
    f^j_{\operatorname{clf}}(\bm{Z}) = a^j_{l_2}\left(\bm{W}^j_2 \cdot\left(a^j_{l_1}\left(\bm{W}^j_{1} \cdot  f^j_{\mathrm{att}}(\bm{Z}) +\bm{b}_{l_1}\right)\right)+\boldsymbol{b}^j_{l_2}\right)\nonumber.
\end{align}
To predict multiple root causes simultaneously in an end-to-end manner, the training cross-entropy loss becomes 
$ \mathcal{L}^{\mathrm{Clf}} =  -\sum_{j} Y_j \log ( \text{softmax}(f_{\operatorname{clf}}^j(\bm{Z}))$.
The total loss for the MMSLA model is the summation of the VAE loss and classification loss:$\mathcal{L} = \mathcal{L}^{\mathrm{VAE}}+ \lambda \mathcal{L}^{\mathrm{Clf}}$,
where $\lambda \ge 0$ is the tuning parameter. 
The model is trained using the standard Adam optimizer, with hyperparameters fine-tuned via cross-validation to optimize generalization performance.

The proposed MMSLA model facilitates precise diagnosis of root causes across the three MII layers. 
To enable mitigation, we adapt established strategies from literature—tailored to data-, AI pipeline-, and cyber-physical -layer hazards. 
A real-world case study (Sec.~\ref{Sec: CaseStudy}) details these mitigations and validates their efficacy through numerical experiments.

% Early detection minimizes downtime by triggering automated mitigations—such as dynamic model retraining for data drift or failover protocols for fog node failures—while accurate diagnosis ensures corrective actions align with the underlying failure mode (e.g., recalibrating sensors versus patching network vulnerabilities). This closed-loop resilience framework transforms disruptions into opportunities for system adaptation, sustaining operational continuity in dynamic manufacturing environments.To enable an AI system to achieve high resilience in MII, it is important to enable the detection of abnormalities and diagnosis of the root cause so that the mitigation actions can be taken correspondingly. 

\section{Case Study}\label{Sec: CaseStudy}
To validate the resilient AI definition, diagnosis, and mitigation framework, we deployed an MII testbed in which the AI system supports the AJP quality modeling as its core computation task.
% To validate the proposed resilience-oriented AI framework, we developed an MII with AJP quality modeling as the supervised learning task supported by the AI system.

\subsection{Experimental Setup}
% introduce based on the AI system, then data, AI-pipeline, cyber-physical layers
We utilize a collected real AJP dataset with a multivariate time series (MTS) classification DNN pipeline \citep{shojaee2021deep}.
The dataset comprises 95 samples, each containing six \textit{in situ} process variables (\ie atomizer gas flow, sheath gas flow, current, nozzle X-coordinate, nozzle Y-coordinate, and nozzle vibration) in time series format, alongside binary quality responses (\ie conforming or nonconforming) determined by the printed circuits' resistance. 
We introduce four synthetic time series variables with negligible predictive power, resulting in ten total inputs for the classification task.

Using a simulation framework, we generate five datasets emulating distinct AJP machines. The AI pipeline (Sec.~\ref{Sec: Intro}) includes data augmentation, standardization, and a DNN classifier, with 128 possible configurations.
The optimal configuration, selected via training on the real dataset, serves as the baseline. 
By perturbing parameters in the baseline’s final layer, we derive five ground-truth models, each reflecting the underlying correlation between the input time series data and output quality response across simulated machines.
These five models provide the ground truth label for the new input samples, which are generated by augmenting the original 95 samples with different methods (\ie timewrapping, pooling, convolving, \etc).
% to mimic the real manufacturing processes by varying the level of different root causes.
Using the new samples and labels, the MTS DNN pipeline is trained for each machine. 
The top three performing pipelines among all configuration for each machine (3 pipelines times 5 machines equals 15 in total) are deployed on computation nodes to predict product-quality during the manufacturing process.
The MII AI system’s computational infrastructure comprises five fog nodes (\eg Raspberry Pis) and a centralized Cloud server, resulting in six computation nodes in total.

% Then about the root cause factors explanation
To simulate the hazards in MII AI systems, we vary the levels of root cause factors from three layers to create different hazard scenarios.
As shown in Table.~\ref{tab:factors}, we introduce the following variations in the data layer:
(i) Sensor contamination and failure are modeled by altering the ratio of affected sensors among ten input variables, replacing signals with constant values, or increasing/decreasing trends to mimic sensor degradation and calibration issues;
(ii) Signal-to-Noise Ratio (SNR) is adjusted by injecting random noise into the input data at varying scales;
(iii) Input distribution shifts are simulated by introducing a mixture of Gaussian-distributed data and controlling the resulting KL divergence from the original distribution;
(iv) Class imbalance is manipulated by augmenting conforming samples and adjusting the nonconforming-to-conforming sample ratio (\ie 40/60, 25/75, 10/90) within each data batch.
For the AI pipeline layer, we simulate pipeline singularity by forcing a subset of pipelines (0 of 3, 1 of 3, or 2 of 3) to produce identical predictions for all data points within a batch with doubled computation time.
For the cyber-physical layer, we simulate communication channel disruption by initiating multiple simultaneous download and upload tasks on a subset of fog nodes (0 of 5, 1 of 5, or 2 of 5), mimicking a DDoS attack. 
Additionally, fog node failure is modeled by disabling a subset of fog nodes, preventing them from receiving or transmitting signals and halting their computation.
\begin{table}[ht] 
\centering
\caption{The root cause factor levels.}
\resizebox{\linewidth}{!}{
\begin{tabular}{llllll}
\toprule
\textbf{\shortstack[l]{Root Cause \\Factors}} & \textbf{Definition} & \textbf{Layer} & \textbf{Level 0} & \textbf{Level 1} & \textbf{Level 2} \\
\midrule
$Y_1$ & \shortstack[l]{\% of Contaminated or \\Failed Sensors} & Data & 0 & 10 & 20 \\
$Y_2$ & SNR & Data & Low & - & High \\
$Y_3$ & \shortstack[l]{Distribution Change of \\ Input Variable} & Data & Nan & Low & High \\
$Y_4$ & Class Imbalancess & Data & 40/60 & 25/75 & 10/90 \\
$Y_5$ & \% of Sinular Pipelines & AI Pipeline & 0 & 1 & 2 \\
$Y_6$ & \shortstack[l]{\% of Failed Egde Nodes} & Cyber-physical & 0 & 1 & 2 \\
$Y_7$ & \shortstack[l]{\% of Failed \\ Communication Channels} & Cyber-physical & 0 & 1 & 2 \\ \bottomrule
\end{tabular}}\label{tab:factors}
\end{table}

In the experiment, each computation task is defined as predicting labels for a batch of data collected from a single machine.
Each task is executed on one of six computation nodes.
Using a full factorial design based on the factors in Table~\ref{tab:factors}, we generate 1,458 normal and hazard scenarios. 
To reflect real manufacturing operations, where hazards are infrequent, we create five computation tasks per scenario, ensuring that at most two tasks per scenario experience hazard conditions (\ie at least one factor is not at Level 0).
This results in a total of 7,290 computation tasks ($1458\times 5$). Each task is randomly assigned to a computation node by the Cloud.
% , which acts as the system orchestrator.

%then about the collected data
After the execution of each computation task, $\bm{X}^P$ are collected as the inference accuracy, precision, and F1 score of the top-3 pipelines on the data batch.
Additionally, $\bm{X}^R$, runtime metrics, are captured at the same registration frequency throughout task execution, which include CPU utilization, CPU temperature, memory consumption, download bandwidth, upload bandwidth, and data transmission volume.

The hyperparameter $\lambda$ is set as 0.1, and $\alpha$ is set as 0.1 by five-fold cross validation (CV). We have $u^R = u^P = 16$.

% # table for collected class distribution?

\subsection{Evaluation of the Diagnosis Model}
% the distribution of the collected data
% evaluation metrics and benchmark methods

We first evaluate the MMSLA model's diagnostic performance, focusing on detecting hazard conditions with one or multiple root causes. 
Assessing the failure severity  level is left for future work.

Since the number of computation tasks under Level 2 and Level 3 for each root cause factor is much smaller than those under Level 0, there exists a class imbalance issue in the diagnosis task. Therefore, the F1 score is chosen as the evaluation metric.
We compare the MMSLA model with three benchmark methods.
The Multimodal Self Latent Attention (MSLA) model is a simplified version of MMSLA, using a single attention head instead of multiple heads. This head is applied to the latent feature $\bm{Z}$ to predict multiple responses directly.
Since no existing methods handle multimodal inputs with varying dimensions, we use Fourier transformation and summary statistics to extract features from $\bm{X}^R$, including sample length, harmonic mean, standard deviation, kurtosis, entropy, second harmonic, third harmonic, and total harmonic distortion.
The extracted features, combined with $\bm{X}^P$, are used as inputs for Random Forest (RF) and XGBoost, two classical supervised learning models designed for high-dimensional complex data.
To enhance the performance of RF and XGBoost, a separate model is trained for each root cause identification. As a result, five models are trained for each of the two benchmark methods.
% Note that one model is trained for the identification of one root cause to improve the performance for Random Forest and XGBoost. There are five models trained for two benchmarks, respectively.
To address the class imbalance, we apply SMOTE to augment the training data \citep{chawla2002smote}.
However, SMOTE is not used for MMSLA and MSLA, as it degrades their performance, which might be due to that neural networks are more sensitive to the quality of augmented data.

We summarize the results for diagnosis factors $Y_1$ to $Y_5$ in Table~\ref{tab:f1_5}. 
$Y_6$ and $Y_7$ are excluded because, under cyber-physical layer failure, the computation task cannot be completed, leading to missing $\bm{X}^P$. As a result, failures caused by these two factors are trivially identifiable and become a binary classification task.
From the results in Table~\ref{tab:f1_5}, the proposed MMSLA model achieves the best performance under four out of five factors.
MSLA performs significantly worse than MMSLA and even underperforms the two benchmark methods for $Y_4$ and $Y_5$.
This may be due to the use of a single attention head, which blends heterogeneous variable dependencies across multiple root causes, limiting its ability to distinguish distinct factors. 
These results validate the effectiveness of the proposed multi-head self latent attention mechanism in capturing root cause-specific dependencies.
In addition, both RF and XGBoost achieve high performance for $Y_4$ and $Y_5$ compared to other factors.
This indicates that distribution change and class imbalance can be relatively easily diagnosed. 
% Both Random Forest and XGBoost achieve high performance for $Y_4$ and $Y_5$  compared to other factors, suggesting that distribution shifts and class imbalance are relatively easier to diagnose.
\begin{table}[t!] 
\centering
\caption{The F1 score of root cause diagnosis for factor $Y_1 - Y_5$. Mean and standard
deviation are reported over 5-fold CV.}
\resizebox{0.8\linewidth}{!}{
\begin{tabular}{cccccc}
\toprule
Method               & $Y_1$           & $Y_2$           & $Y_3$           & $Y_4$           & $Y_5$           \\
\midrule
MMSLA (Proposed)                & \textbf{0.90} (0.01) & \textbf{0.70} (0.02) & \textbf{0.89} (0.01) & 0.94 (0.01) & \textbf{0.95} (0.01) \\
MSLA                 & 0.86 (0.02) & 0.61 (0.01) & 0.89 (0.01) & 0.71 (0.01) & 0.76 (0.02) \\
\shortstack{XGBoost + \\SMOTE}      & 0.79 (0.00) & 0.68 (0.01) & 0.78 (0.01) & 0.94 (0.01) & 0.89 (0.00) \\
\shortstack{Random Forest+ \\SMOTE} & 0.62 (0.01) & 0.56 (0.01) & 0.70 (0.01) & \textbf{0.95} (0.02) & \textbf{0.95} (0.01)\\
 \bottomrule
\end{tabular}}\label{tab:f1_5}
\end{table}

For the classification between factor $Y_6$ and $Y_7$ where $\bm{X}^P$ is used as the input, both RF and XGBoost with SMOTE achieve satisfactory performance as shown in Table~\ref{tab:f1_2}.
\begin{table}[t!] 
\centering
\caption{The F1 score of root cause diagnosis for factor $Y_6, Y_7$. Mean and standard
deviation are reported over 5-fold CV.}
\resizebox{0.5\linewidth}{!}{
\begin{tabular}{ccc}
\toprule
Method               & $Y_6$           & $Y_7$          \\
\midrule
XGBoost + SMOTE     & 0.90 (0.00) & 0.89 (0.01) \\
Random Forest+ SMOTE & 0.92 (0.01) & 0.91 (0.01) \\
 \bottomrule
\end{tabular}}\label{tab:f1_2}
\end{table}
% Random Forest slightly outperforms XGBoost, achieving an average F1 score of 0.92 with a standard error of 0.01. Yingyan: Would it be better just list two columns of Y6 and Y7 for F1 scores?

\subsection{Resilience Metrics and Mitigation Strategies} % ad-hoc from literature
% explain one of the three mitigation strategies and demonstrate the numerical example
Building on the diagnostic model’s outputs, we adapt context-aware mitigation strategies from literature tailored to the diagnosed hazard type (data-, AI pipeline-, or cyber-physical-layer), and quantify system recovery using the temporal and performance resilience metrics defined in Section~\ref{Sec: methodology}.

\begin{figure}[h!]
    \centering
    \includegraphics[width=0.8\linewidth]{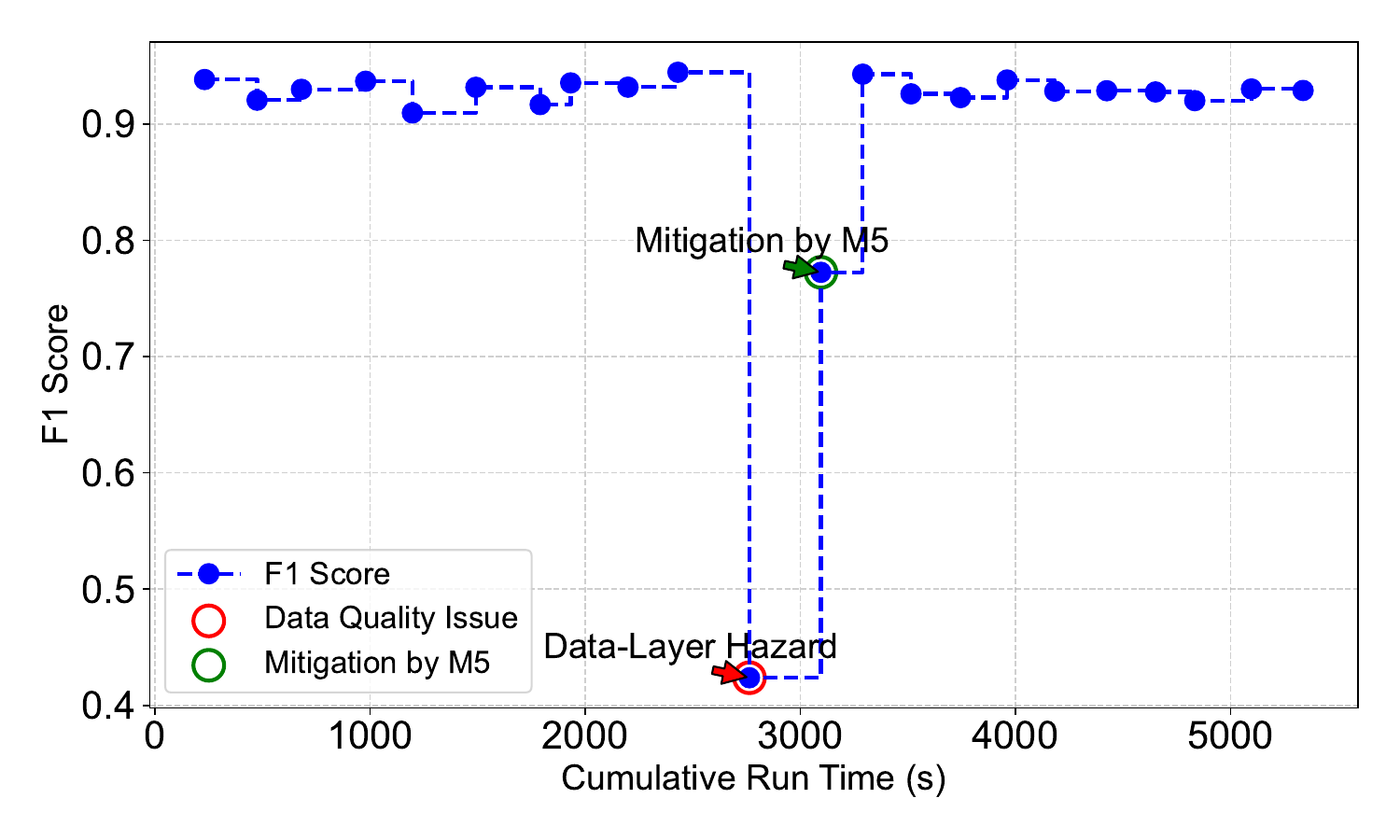}
    \caption{AI system performance under data hazards and mitigation.}
    \label{fig:quality_miti}
\end{figure}

For data-layer hazards, we implement automated data substitution, replacing corrupted inputs with data from a similar process \citep{whang2023data}. 
This immediate mitigation stabilizes system performance, while root-cause-specific protocols, such as sensor recalibration for SNR degradation or synthetic oversampling for class imbalance, are initiated to address the diagnosed failure mode.
In particular, the data generated for the same scenario for another machine but without hazards is selected, where the machine is selected by the largest cosine similarity between the last layer parameters and the machine that generates the data with quality issues.
Fig.~\ref{fig:quality_miti} visualizes the AI system performance on a computation node under data layer hazards and the effect of the proposed mitigation strategy. 
A sequence of computation tasks is assigned to the node, with the X-axis representing the cumulative execution time and the Y-axis showing the F1 score of the product quality prediction task from the best result among the three deployed pipelines, reflecting the system’s performance.
The diagnosis model is applied to monitor the system.
The red point represents the task diagnosed with data-layer hazards, where the data are generated from Machine 1 (M1). 
After substituting the data source with M5, the performance recovers to the green point.
Let $P_S = 0.75$ and $\Delta t = 400$ s as the average duration for each task is $360$ s, the temporal and performance resilience can be calculated as: $\text{FD} = 664.58,  \text{PR} = 0.598, \text{RR} = 0.822$.
Since performance is recorded only upon task completion, the step function results in $t_1 = t_2$. As a result, recovery efficiency is not computed in this case.
% Other in-processing methods such as data cleaning
% adopt the change the data source from a 
% Random order the computation tasks
% Online monitoring

\begin{figure}[h!]
    \centering
    \includegraphics[width=0.8\linewidth]{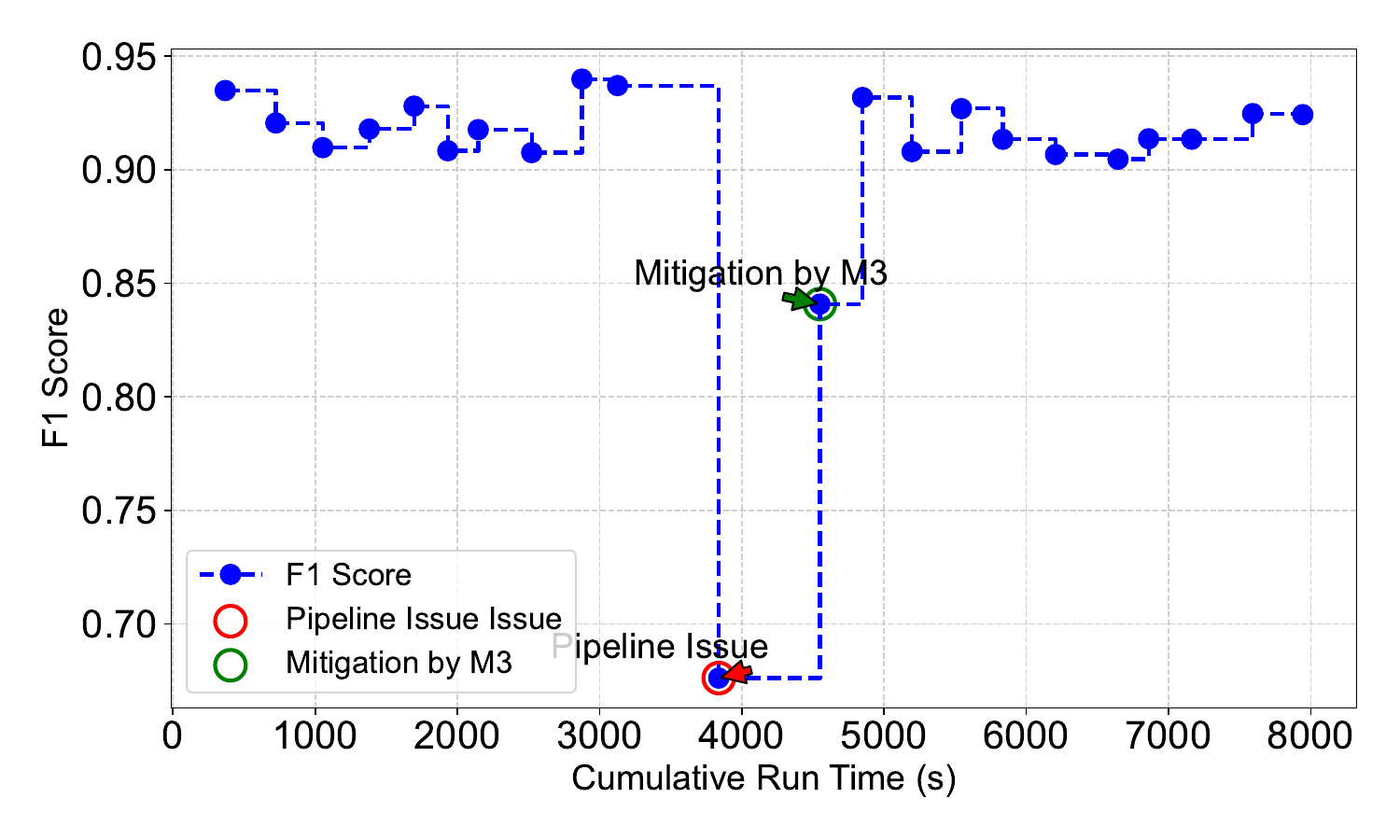}
    \caption{AI system performance under pipeline hazards and mitigation.}
    \label{fig:pipe_miti}
\end{figure}

For the hazards in the AI pipeline layer, we replace the pipeline with the ones from a similar machine (\ie the machine with the largest cosine similarity between the last layer parameters and the machine that faces pipeline singularity hazard).
Similarly, Fig.~\ref{fig:pipe_miti} illustrates the MII testbed system performance on a computation node under pipeline layer hazards and the impact of replacing the pipelines with those from Machine 3 (M3).
The corresponding temporal and performance resilience can be calculated as: $\text{FD} =711.39, \text{PR} = 0.759, \text{RR} = 0.897 $.
Comparing PR and RR between the two mitigation actions in Fig.~\ref{fig:quality_miti} and Fig.~\ref{fig:pipe_miti} reveals that the hazard in Fig.~\ref{fig:quality_miti} is more severe, causing a greater impact on the AI system, as indicated by its lower PR and RR.

For cyber-physical layer hazards, when a fog node or communication channel failure is detected, the Cloud orchestrator dynamically reassigns the affected computation task to an available node, minimizing disruption and maintaining system resilience. The demonstration is omitted here due to the change in computation nodes.

The empirical results validate the effectiveness of mitigation strategies upon diagnosing the root causes from the literature and demonstrate the utility of the proposed resilience metrics in quantifying the AI system's ability to withstand and recover from hazards in MII.

\section{Conclusion and Future Work}\label{Sec:conclusion}

With the growing adoption of AI systems in MII for critical decision-making, ensuring AI resilience has become a significant challenge for cybermanufacturing operations. However, there was a knowledge gap in defining AI resilience, identifying root causes of failures, and developing effective mitigation strategies.
In this work, we propose a resilient AI framework for MII, focusing on the AI pipeline for online prediction tasks. We analyze AI performance under hazards from the data layer, AI pipeline layer, and cyber-physical layer, introducing temporal and performance resilience metrics to quantify system resilience. To enable accurate failure diagnosis, we develop the MMSLA model, which effectively captures dependencies within multimodal data of varying dimensions and accurately identifies specific root causes. Additionally, we propose layer-specific mitigation strategies to enhance system robustness.
The effectiveness of the proposed resilient AI framework is validated through an MII testbed, where the MMSLA model outperforms multiple benchmark methods in diagnosis accuracy.
As future work, we aim to develop a higher-resolution diagnosis model capable of detecting hazard severity levels and explore online mitigation strategies to further enhance AI system resilience in MII by improving data quality \citep{zeng2023ensemble}, AI pipeline uncertainty quantification and ranking \citep{chen2024lori}, and computation offloading \citep{chen2018predictive}.
% , nallendran2021predictive
% With the growing adoption of AI systems in MII for critical decision-making, the resilience of AI systems has posed significant challenges to manufacturing operations.
% There exists a wide knowledge gap in defining AI resilience, identifying the potential root causes, and providing mitigation strategies.
% In this work, we propose a resilient AI framework in MII, focusing on the AI pipeline for online prediction tasks.
% The AI performance is investigated under hazards in the data, AI pipeline, and cyber-physical layer in the system, the corresponding temporal and performance resilience metrics are proposed to quantify the system resilience.
% To isolating failures for accurate diagnosis, we propose a MMSLA model which learns the dependencies between multimodal data with varying dimensions and effectively specific root causes effectively.
% Furthermore,  mitigation strategies for failure in each layer are proposed. 
% The effectiveness of the resilient AI framework is validated within an MII testbed.
% The MMSLA model is benchmarked against multiple existing methods and demonstrates superior performance.
% As future work, we will investigate diagnosis model with higher resolution which can diagnose the level of each hazard, and also investigate online mitigation strategies to make the AI system with higher resilience.
% \section{Acknowledgement}
% \vspace{-0.4em}
% The authors acknowledge the financial support from XXX.

\bibliographystyle{plainnat}
\bibliography{Ref.bib}
% \bibliography{Ref.bib}

\end{document}